\documentclass[conference, 9pt]{IEEEtran}
\IEEEoverridecommandlockouts
\usepackage{cite}
\usepackage{times}
\usepackage{amsmath,amssymb,amsfonts}
\usepackage{algorithmic}
\usepackage{graphicx}
\usepackage{textcomp}
\usepackage{xcolor}
\def\BibTeX{{\rm B\kern-.05em{\sc i\kern-.025em b}\kern-.08em
    T\kern-.1667em\lower.7ex\hbox{E}\kern-.125emX}}

\newcommand{\OURS}{TACOS}
\usepackage{booktabs}
\usepackage[accsupp]{axessibility}
\usepackage{array}    % 用于增强表格功能
\usepackage{arydshln} % 用于虚线
\usepackage{booktabs}
\usepackage{colortbl} % for row colors
\usepackage{xcolor}

\usepackage{floatpag}
\usepackage{multirow}
\usepackage{tabularx,ragged2e,caption}
\usepackage{adjustbox} % 调整盒子大小的宏包
\usepackage{lipsum}    % 用于生成示例文本
\usepackage{etoolbox}
\usepackage{url}
\usepackage[normalem]{ulem}

\begin{document}

\title{TACOS: Open Tagging and Comparative Scoring for Instruction Fine-Tuning Data Selection}

\author{
    \IEEEauthorblockN{
        Xixiang He\textsuperscript{1},
        Hao Yu\textsuperscript{2},
        Qiyao Sun\textsuperscript{1}, 
        Ao Cheng\textsuperscript{1}, \\
        Tailai Zhang\textsuperscript{1},
        Cong Liu\textsuperscript{1},
        and Shuxuan Guo\textsuperscript{2*}
        \thanks{\textsuperscript{*} denotes the corresponding author.}
    } \\
    \IEEEauthorblockA{\textsuperscript{1}National University of Defense Technology, Changsha, China} 
    \IEEEauthorblockA{\textsuperscript{2}Intelligent Game and Decision Lab, Beijing, China} 
    \IEEEauthorblockA{\{hexixiang, sunqiyao18, chengao18, zhangtailai, liucong21\}@nudt.edu.cn, \\haoyutum@163.com, sg10.work@outlook.com}
}

\maketitle

% \author{\IEEEauthorblockN{1\textsuperscript{st} Xixiang He}
% \IEEEauthorblockA{\textit{National University of Defense Technology} \\
% ChangSha, China \\
% hexixiang@nudt.edu.cn}
% \and
% \IEEEauthorblockN{2\textsuperscript{nd} Hao Yu}
% \IEEEauthorblockA{\textit{Intelligent Game and Decision Lab} \\
% Beijing, China \\
% haoyutum@163.com}
% \and
% \IEEEauthorblockN{3\textsuperscript{rd} Qiyao Sun}
% \IEEEauthorblockA{\textit{National University of Defense Technology} \\
% ChangSha, China \\
% sunqiyao18@nudt.edu.cn}
% \and
% \IEEEauthorblockN{4\textsuperscript{th} Ao Cheng}
% \IEEEauthorblockA{\textit{National University of Defense Technology} \\
% ChangSha, China \\
% chengao18@nudt.edu.cn}
% \and
% \IEEEauthorblockN{5\textsuperscript{th} Tailai Zhang}
% \IEEEauthorblockA{\textit{National University of Defense Technology} \\
% ChangSha, China \\
% zhangtailai@nudt.edu.cn}
% \and
% \IEEEauthorblockN{6\textsuperscript{th} Cong Liu}
% \IEEEauthorblockA{\textit{National University of Defense Technology} \\
% ChangSha, China \\
% liucong21@nudt.edu.cn}
% \and
% \IEEEauthorblockN{7\textsuperscript{th} Shuxuan Guo}
% \IEEEauthorblockA{\textit{Intelligent Game and Decision Lab} \\
% Beijing, China \\
% sg10.work@outlook.com}

% }

\begin{abstract}
Instruction Fine-Tuning (IFT) is crucial for aligning large language models (LLMs) with human preferences, and selecting a small yet representative subset from massive data significantly facilitates IFT in terms of both efficiency and effectiveness. Nevertheless, existing approaches suffer from two limitations: the use of simple heuristics restricts data diversity, while the singleton data quality evaluation accounts for inconsistent criteria between independent samples. 
To address the issues, we present~\OURS, an innovative method that integrates Open Tagging and Comparative Scoring for IFT data selection. 
To capture data diversity, we leverage LLMs to assign open-domain tags to human queries, followed by a normalization stage to denoise the open tags and enable efficient clustering.
Additionally, we suggest a comparative scoring method that allows the relative quality evaluation of samples within a cluster, avoiding inconsistent criteria seen in singleton-based evaluations.
Extensive experiments across diverse datasets and LLM architectures demonstrate that~\OURS~outperforms existing approaches by a large margin.
Notably, it achieves superior instruction-following performance on MT-Bench and ranks 1st among LLaMA2-7B-Based models on AlpacaEval 2.0, illustrating its efficacy for IFT data selection.
\end{abstract}
%~\footnote{\url{https://github.com/hexixiang/TACOS}} Codes are available at \url{https://github.com/hexixiang/TACOS}.

\section{Introduction}
\label{sec:intro}

However, direct IFT with unfiltered massive data accounts for flaws such as inefficient use of computational resources, increased time cost, contamination from noisy data, etc. 
Consequently, the emphasis has moved towards selecting a small subset of high-quality data for IFT, which enhances LLM performance more effectively~\cite{ft1}.

Although the research field of IFT data selection has witnessed rapid development in recent years, existing works~\cite{Lima,chen2024alpagasus,chen2023improving,what,longest,IFD,ppl,Du2023MoDSMD,mukherjee2023orca,zhao2023preliminary,longpre2023flan, kumar2019submodular,li2022consisttl} are constrained by two major challenges: preserving data diversity after selection and evaluating data quality with consistent criteria.
To enhance data diversity, previous approaches mainly rely on intuitive human prior, such as selecting instruction pairs with the longest response~\cite{longest} or leveraging linear quality rules~\cite{Cao2023InstructionMH}. 
\begin{figure}[t]
    \centering
    \includegraphics[width=1.0\columnwidth]{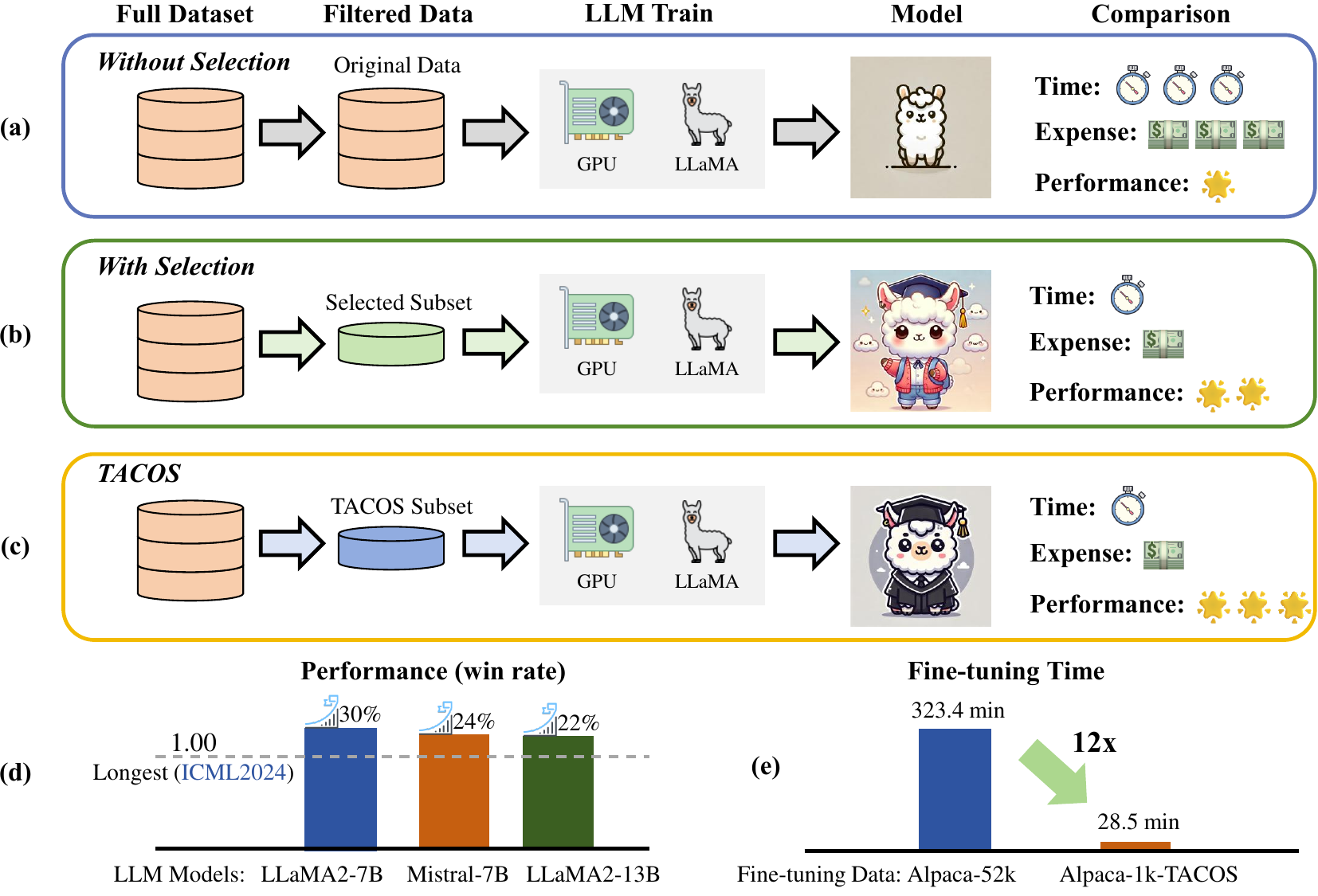}
    \caption{\textbf{Top:}~Comparison of different IFT data selection strategies, including: \textbf{(a)} original data based IFT, which is time-consuming and expensive, while leading to suboptimal performance; \textbf{(b)} selected data based IFT, which saves training time and expense, while improving IFT performance; \textbf{(c)~\OURS} that introduces \textit{Open Tagging} and \textit{Comparative Scoring} to further boost the performance of data selection for IFT. \textbf{Bottom:} Quantitative results, including: \textbf{(d)} comparison between \OURS~and a SOTA baseline for IFT data selection in terms of LLM performance, where our consistent higher win rate on variant LLMs demonstrates our superiority. \textbf{(e)}~comparison between \OURS~based and original data based IFT in terms of LLM fine-tuning time, where \OURS~achieves 12x acceleration.}
    \label{fig:tiser}
    \vspace{-0.8cm}
\end{figure}

\begin{figure*}[htbp] % 使用 figure* 环境来跨栏显示，并指定 [t] 选项放在页顶
    \centering
    \includegraphics[width=\textwidth]{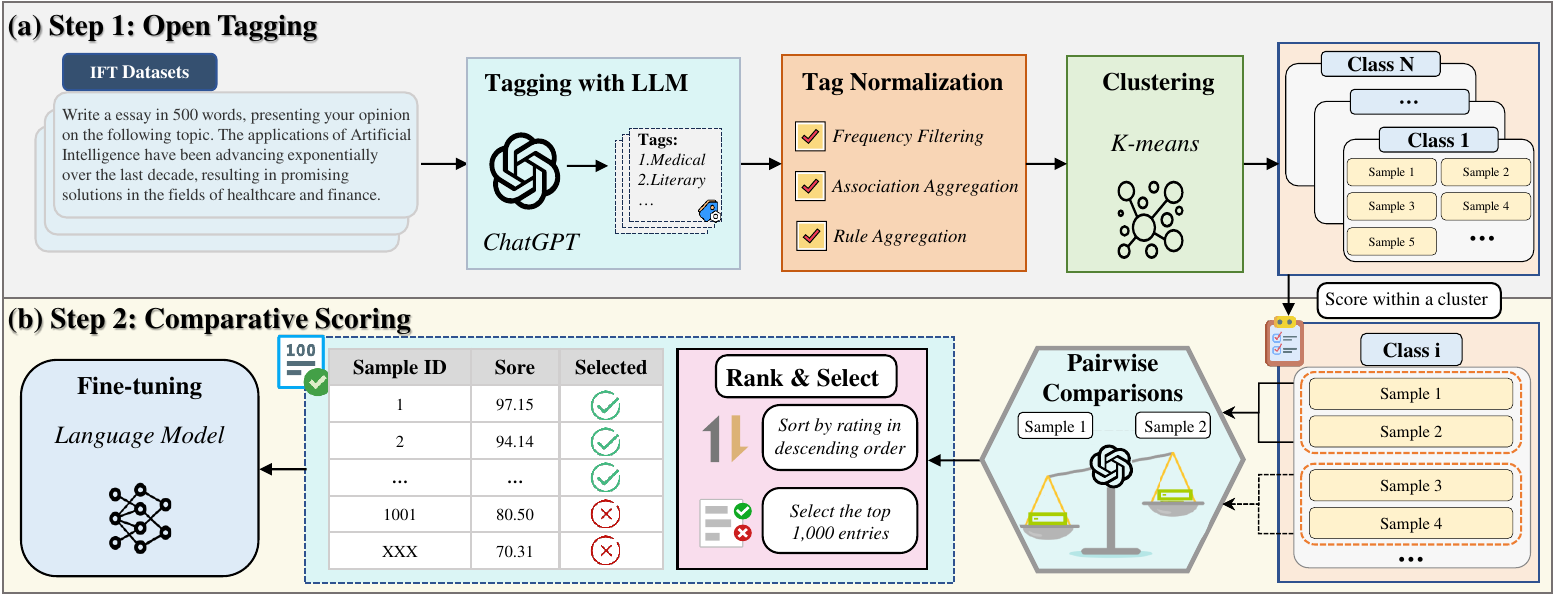}
    \caption{\textbf{An overview of~\OURS}. \textbf{Top:} (a)~\textbf{Open Tagging}. From left to right, a LLM is leveraged to generate open-domain tags for IFT datasets, followed by normalization and clustering to group similar samples, ensuring data diversity and efficiency simultaneously. \textbf{Bottom:} (b) \textbf{Comparative Scoring}. From right to left, a LLM is used to perform comparative scoring within each cluster to obtain consistent criteria and reliable scores for high-quality IFT data selection.}
    \label{fig:overview}
\end{figure*}

% Nevertheless, those simple heuristics fail to capture the semantic level diversity encoded in textual content and therefore lead to suboptimal data diversity preservation.
Nevertheless, those simple heuristics fail to capture the semantic level diversity encoded in textual content, leading to suboptimal data diversity preservation. On the other hand, to set criteria for data evaluation, recent works~\cite{chen2024alpagasus,chen2023improving,what} have proposed to leverage learned prior knowledge as constrains through rating each instruction-response pair by LLMs. However, due to the absence of reference when evaluating each individual data sample, the criteria remain inconsistent, i.e., high-quality data is likely to be assigned with low scores, leading to unreliable data ranking in IFT data selection.

To tackle these issues, we introduce a novel IFT data selection method, termed~\OURS~(Open \textbf{TA}gging and \textbf{CO}parative \textbf{S}coring), which consists of two primary modules: (1) \textbf{Open Tagging}, where IFT data is first annotated with open-domain tags created by LLMs for preserving data diversity. Normalization is further introduced to filter out redundant tags, facilitating a more efficient clustering afterwards. (2) \textbf{Comparative Scoring}, where prompts are first refined to align LLM evaluators with human experts. LLMs are then adopted to perform comparative scoring, i.e., taking another data sample as reference when evaluating one data sample, which helps keep consistent evaluation criteria across all the data and enhances the stability of the consecutive score-based data selection. To ensure consistency and mitigate LLM bias, we swap and rescore the evaluated and reference samples. Benefiting from both designs,~\OURS~consistently surpasses baseline methods for the IFT of LLMs. Furthermore, we only use publicly available data for training and have released both the code and data for reproducibility~\footnote{\url{https://github.com/hexixiang/TACOS}}. 

Our main contributions are summarized as follows:
\vspace{0.4em}
\begin{itemize}
    \item We propose to introduce open tagging with normalization procedures, ensuring data diversity and clustering efficiency.
    \vspace{0.3em}
    \item We design to refine data evaluation prompts and evaluate data quality in a comparative fashion, thereby enhancing the reliability and consistency of assessments for IFT data selection.
    \vspace{0.3em}
    \item We conduct extensive experiments on a variety of LLMs, IFT datasets, and benchmarks, demonstrating the efficiency and effectiveness of~\OURS~for IFT data selection.
\end{itemize}

\section{RELATED WORK}
\subsection{Instruction Fine-Tuning of LLMs}
IFT has emerged as a critical step to bridge the gap between LLMs' predictive objectives and user goals, improving response controllability and computational efficiency~\cite{ft2, ft3}. 
Notable models (e.g., InstructGPT~\cite{instructgpt}) employ a multi-step approach combining IFT and reinforcement learning with human feedback (RLHF)~\cite{kirk2023understanding}, which enhances model accuracy, utility, and safety. Moreover, specialized IFT approaches, such as Alpaca~\cite{StanfordAlpaca}, Vicuna~\cite{chiang2023vicuna} and ChatGLM4~\cite{glm2024chatglm}, have demonstrated effectiveness in domain-specific or multilingual contexts, improving instruction adherence and reducing potential biases. Consequently, IFT remains a powerful strategy for advancing LLM capabilities across a wide range of tasks.

\subsection{Data Selection for Instruction Fine-Tuning}
LLMs have demonstrated notable capabilities across diverse tasks, but the high time and computational costs associated with IFT constrain their scalability and limit broader real-world adoption. 
Some studies~\cite{Lima, touvron2023llama, wang2022super} suggest that artificially constructing IFT datasets with diverse instructions and uniform response styles can yield strong fine-tuning results. Yet, there is currently no unified paradigm for how to automatically select and construct such datasets, prompting research into more effective IFT data selection strategies.
Early metric-based approaches rely on simple indicators, such as instruction length~\cite{longest}, Instruction-Following Distance (IFD)~\cite{IFD}, or perplexity~\cite{ppl} to gauge data quality. While these methods are straightforward, they overlook nuanced contextual relationships among data samples. 
% As pioneers, indicator-based techniques utilize simple natural language metrics, such as instruction length~\cite{longest}, IFD~\cite{IFD} and perplexity~\cite{ppl}, to estimate IFT data quality and relevance but often miss contextual relationships between samples.
In contrast, model-based solutions employ fully trained LLMs to score and filter data, resulting in more detailed assessments but incurring high resource costs. Extending this idea, some advanced LLM-based methods~\cite{chen2024alpagasus} refine this idea by systematically using prompts to evaluate instruction quality, although they risk overfitting to model-specific biases. Although comparative strategies such as winning rates and internal or external comparisons~\cite{kung2023active} help to gauge efficacy, challenges remain to ensure data diversity, stable scoring, and reliable performance across varied domains.
% Secondly, some approaches leverage LLMs themselves to act as automated scorers, effectively identifying high-quality data by assigning scores based on predefined criteria. Finally, more advanced methods involve training specialized models to screen datasets. These models assess data samples using complex scoring mechanisms, often combining quality, diversity, and task-specific relevance. 
% Existing methods for data selection can be broadly categorized into metric-based, model-based, strong LLM-based, and small-model-based approaches. Metric-based methods use predefined criteria like instruction length or perplexity to rank data but often miss contextual relationships between samples. Model-based methods assess data quality using trainable LLMs, offering a more dynamic evaluation but at the cost of scalability. Strong LLM-based methods, such as those leveraging models like ChatGPT, evaluate instruction quality through specially designed prompts, though they risk overfitting to model biases. Small-model-based approaches, utilizing lightweight models or embeddings, provide scalability but may lack the nuance captured by larger models. To assess the effectiveness of these methods, comparison strategies like winning rates and internal/external comparisons are used, although challenges remain in ensuring data diversity and stable, reliable scoring.

\section{METHOD}
As shown in Fig.~\ref{fig:overview},~\OURS~has two key modules: \textbf{Open Tagging}~(\ref{sec:sec2.1}) and \textbf{Comparative Scoring}~(\ref{sec:sec2.2}) that enhance IFT data selection by ensuring data diversity and score reliability, respectively.

\subsection{Open Tagging} \label{sec:sec2.1}
\begin{figure}[htbp]
    \centering
    \includegraphics[width=1.0\columnwidth]{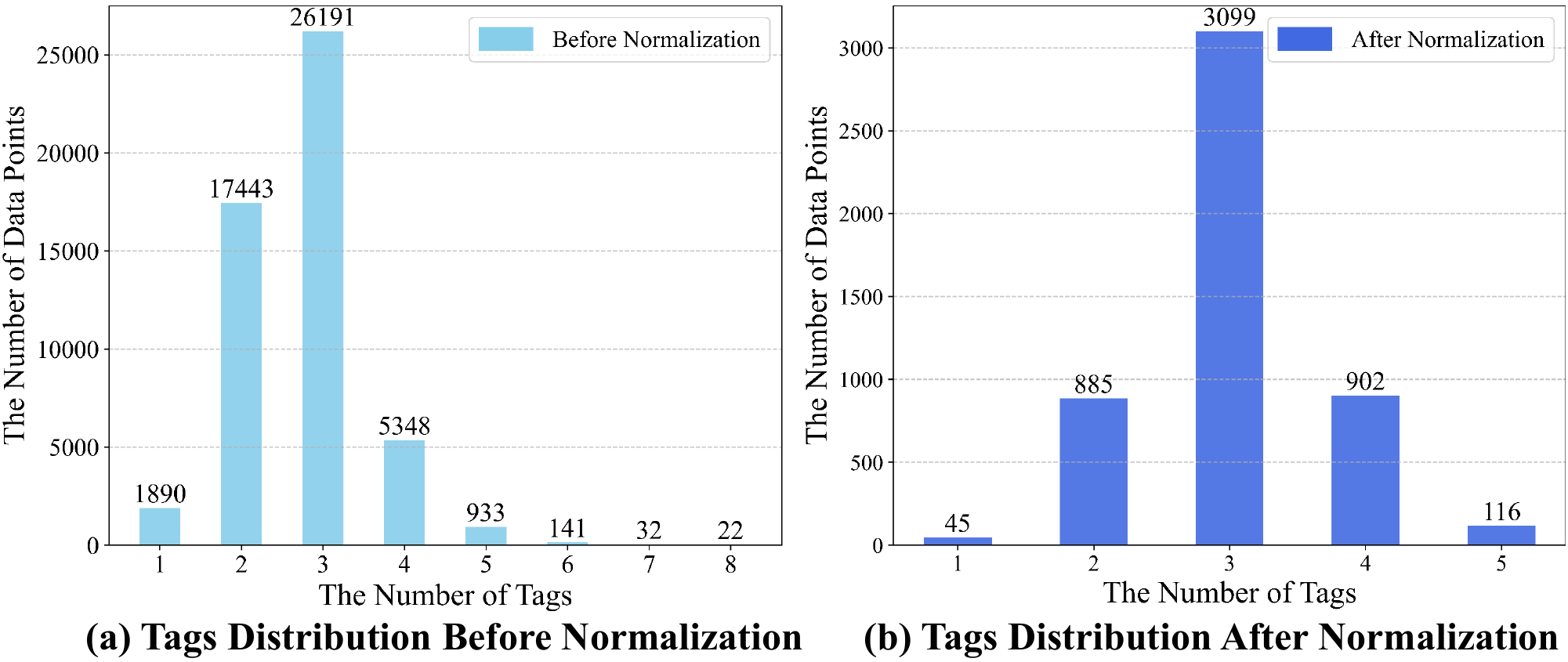}
    \caption{\textbf{Distribution of Tags before and after Normalization.} The introduced normalization procedures compress the size of the original tag set from around 50k to less than 6k.}
    \label{fig:tag_count}
\end{figure}

\noindent\textbf{Open-Domain Tagging.} To capture diverse intentions in IFT data, we propose to eliminate the inherent limitation of pre-defined tags~\cite{instag} by leaving the tag space open to LLMs. To achieve that, we extend the prompt designed in~\cite{instag} and employ GPT-4o~\footnote{\url{https://platform.openai.com/docs/models/gpt-4o}} to annotate instruction-response pairs with fine-grained intentions in a specific JSON format. Through this fashion, GPT-4o generates over 50k tags for Alpaca-52k~\cite{StanfordAlpaca} (as shown in Fig. \ref{fig:tag_count}.(a)), demonstrating LLMs' capability for breaking down the barriers introduced by human prior knowledge. However, letting LLMs decide the tag space freely introduces additional noise, such as non-compliant JSON formats, inconsistent word expressions, and uneven tag granularity.
More importantly, the rapidly increasing number of tags has a significant impact on clustering's computational efficiency.
To address these issues, we introduce normalization procedures~\cite{instag}, including frequency filtering that removes long-tail tags, association aggregation that mines tag relationships, and rule aggregation~\cite{han2000mining} that manually integrates tags. As a result, low-frequency tags are eliminated, and JSON formats are standardized. As shown in Fig.~\ref{fig:tag_count}, these steps significantly reduce tag redundancy and noise, and meanwhile maintain data diversity.

\noindent\textbf{Post-Tagging Clustering.} Different from~\cite{Chen2023MaybeO0} that clusters data based on a combination of instructions and input texts, we use tags for clustering, as tags help identify different tasks and cover a more diverse dataset. Thereby, we propose to utilize normalized tags associated with instruction-response pairs for clustering. To reduce redundancy, we generate semantic vectors for tags using the Phrase-BERT model~\cite{wang2021phrase} and cluster them via unsupervised clustering~\cite{k-means}. When a single instance has multiple tags mapped to different clusters, we unify or reassign them if they frequently co-occur or are semantically similar, thereby resolving conflicts.
After clustering, representative labels of the clusters replace the original tags, significantly reducing redundancy. We then organize these labels into semantically coherent groups, using the longest instruction in each group as a representation for creating a streamlined IFT data subset. In doing so, massive raw data are compressed into a representative subset with the diversity preserved, which significantly facilitates the IFT procedure in terms of both efficacy and efficiency.

% they offer more concise semantic signals to distingish different tasks. Specifically, we embed the tags using the Phrase-BERT model~\cite{wang2021phrase}—which excels at short-phrase similarity—and then apply unsupervised clustering~\cite{k-means}. Before clustering, we remove long-tail tags through frequency filtering and unify inconsistent tag expressions via rule-based normalization, thus reducing noise while retaining sufficient diversity. Afterward, each instruction-response pair inherits the cluster labels of its tags, and we select the longest instruction in each cluster to represent that cluster’s semantics when constructing the final IFT data subset. This strategy not only reduces redundancy but also preserves the coverage of various tasks, ensuring that each cluster remains semantically distinct in the streamlined data for fine-tuning.}~\sout{For computational efficiency, we further propose to utilize normalized tags associated to instruction-response pairs for clustering.
% To reduce redundancy, we generate semantic vectors for tags using the Phrase-BERT model~\cite{wang2021phrase} and cluster them via unsupervised clustering~\cite{k-means}.
% ~\sout{We then assign these tags to semantically diverse clusters, using the longest instruction in each cluster as the representation for creating a streamlined IFT data subset. By doing that, the IFT subset consists of data samples from each semantically diverse cluster, which guarantees data diversity after IFT data selection.}

\begin{figure}[htbp]
    \centering
    \includegraphics[width=1.0\columnwidth]{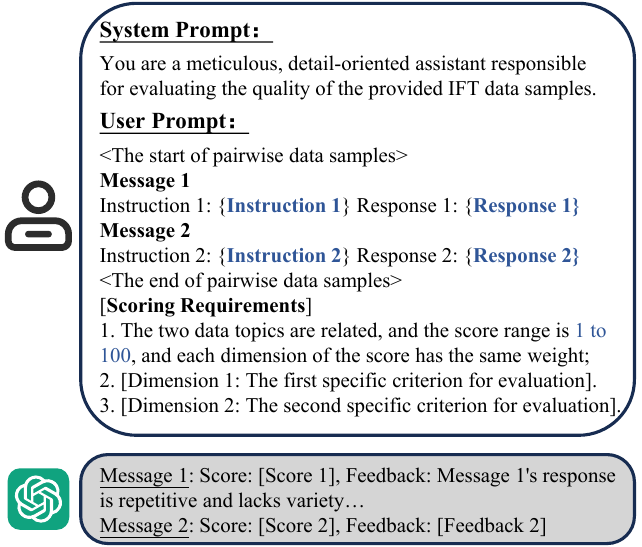}
    \caption{\textbf{Template of Comparative Scoring Prompt.} The LLM evaluates pairwise data in a comparative fashion, scoring each on a range of [1, 100]. Feedback ensures accurate assessment.}
    \label{fig:prompt}
\end{figure}

\subsection{Comparative Scoring} \label{sec:sec2.2}
\noindent\textbf{Prompt Refinement.} Unlike existing works~\cite{chen2024alpagasus,what,wu2023self,kung2023active} that primarily adopt LLMs for data scoring without prompt customization, we argue that careful prompt design, aligned with human evaluation standards, greatly enhances the precision and stability of IFT data selection.
Based on a similar idea, Chen et al.~\cite{chen2023improving} have designed prompts for LLMs to score IFT data responses.
Nevertheless, their narrow score range and lack of human-machine alignment lead to inflated scores as well as gaps between LLM and human assessments. To address this, we refine the prompt by expanding the score range to [1, 100] and adding criteria in the [\textbf{scoring requirements}] part~(see the upper part of Fig.~\ref{fig:prompt}). Such a design aligns scoring standards with human experts better and allows LLMs to provide more stable scores to data samples, improving the accuracy and stability of data evaluation.

\noindent\textbf{Pairwise Scoring.} Previous approaches~\cite{chen2024alpagasus, what, chen2023improving} leverage LLMs to rate each data sample individually and consecutively filter data based on a defined threshold, resulting in unreliable data evaluation and selection. 
To tackle the problem, we propose to cluster all the data first and then leverage GPT-4~\cite{achiam2023gpt} to perform comparative scoring by comparing data samples in a pairwise fashion.
Data with the highest scores is finally selected for IFT. As our method considers mutual relationships by comparing pairs of data samples for scoring, it reduces rating biases and score inflation caused by inconsistent scoring criteria in previous methods, thereby improving the reliability and effectiveness of data selection.

\section{MAIN RESULTS}
\subsection{Experiment Settings}
We fine-tune LLaMA2-7B~\cite{touvron2023llama}, LLaMA2-13B~\cite{touvron2023llama}, and Mistral-7B-v0.1 (abbreviated as Mistral-7B)~\cite{jiang2023mistral} on IFT datasets Alpaca-52k~\cite{StanfordAlpaca} and Evol-Instruct-70k~\cite{xu2023wizardlm} using LLaMA-Factory~\cite{llamafactory} in bfloat16 on 8*A800 GPUs. For the selected 1k IFT data, we adopt the hyperparameters from~\cite{longest, Lima}, with a sequence length of 2,048, a batch size of 128, and gradient accumulation. Models are trained for 3 epochs on the original Alpaca and Evol-Instruct datasets, and 15 epochs on the selected 1k datasets. Five datasets, including LIMA~\cite{Lima}, Vicuna~\cite{chiang2023vicuna}, Koala~\cite{koala_blogpost_2023}, WizardLM~\cite{xu2023wizardlm}, and Self-Instruct~\cite{wang-etal-2023-self-instruct} , are used for testing. Pairwise comparisons based on GPT-4 are used to judge win rates, with ties being allowed. Additionally, models are tested on MT-Bench~\cite{mt-bench} to assess instruction-following abilities across a range of tasks.
\begin{figure}[htbp]
    \centering
    \includegraphics[width=1.0\columnwidth]{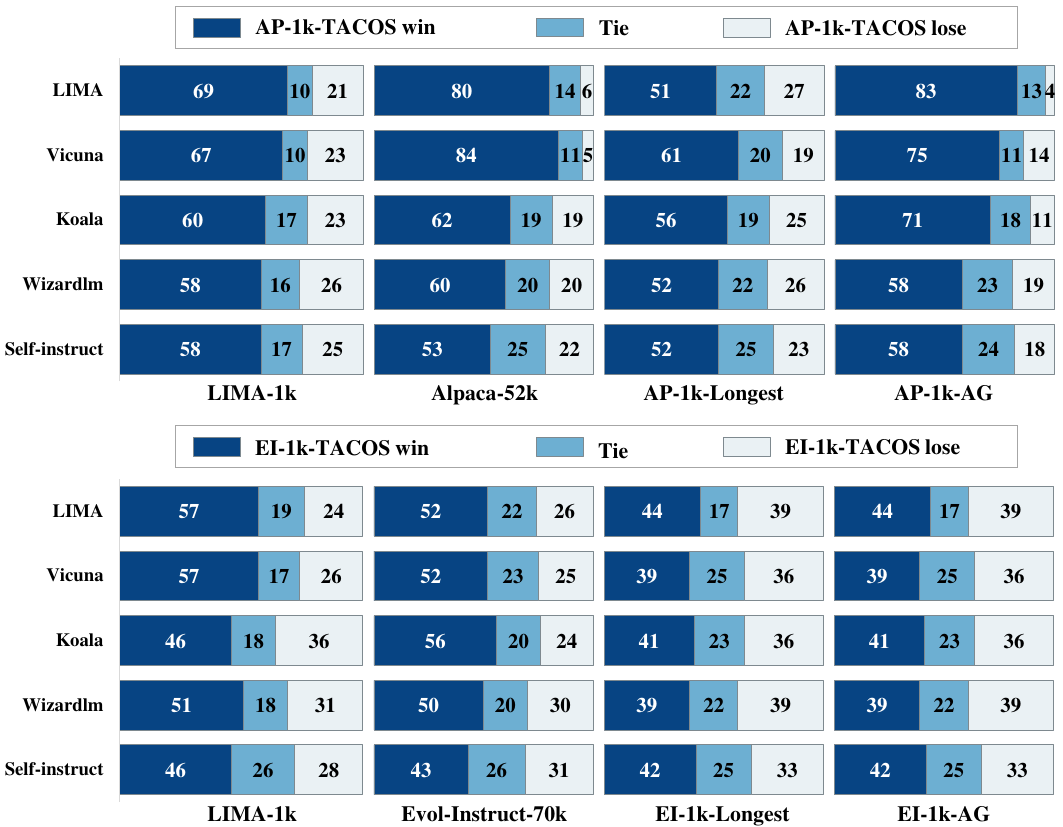}
    \caption{\textbf{Preference evaluation Results~(in~\!\%).} Rows represent five test sets, and columns stand for four baseline methods. The results present the win, tie, and lose rates of~\OURS~versus baselines. \textbf{Top}: comparisons on the Alpaca-52k dataset with LLaMA2-7B. \textbf{Bottom}: comparisons on the Evol-Instruct-70k dataset with LLaMA2-7B. Results demonstrate that our approach consistently yields higher preference scores compared to existing methods.}
    \label{fig:main_experiment}
\end{figure}

\subsection{Head-to-Head Comparisons}
To substantiate the robustness and precision of our data selection strategy in diverse evaluation contexts, we use LLaMA2-7B as the base LLM and apply our method to Alpaca-52k and Evol-Instruct-70k to select 1k data samples, obtaining subsets termed AP-1k-\OURS~and EI-1k-\OURS, respectively. Several baselines with the same base LLM are used for comparisons, including full datasets~(Alpaca-52k and Evol-Instruct-70k), the 1k datasets selected by Zhao et al.~\cite{longest} (namely AP-1k-Longest and EI-1k-Longest), the 1k datasets with the highest scores according to GPT-3.5-Turbo as done by Chen et al.~\cite{chen2024alpagasus} (named AP-1k-AG and EI-1k-AG), and LIMA-1k which is manually curated by Zhou et al.~\cite{Lima}

\begin{figure}[htbp]
    \centering
    \includegraphics[width=1.0\columnwidth]{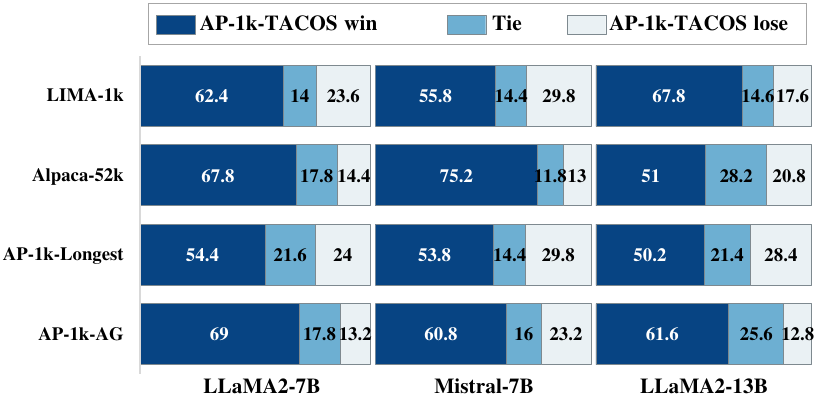}
    \caption{\textbf{Cross-LLM performance~(in~\!\%)}. Rows represent four baseline models, and columns stand for three LLM architectures. The reported results present competition between~\OURS~and
baselines averaged on five evaluation sets. Results illustrate that~\OURS~consistently outperforms baselines on different LLMs, demonstrating~\OURS's excellent instruction-following capability across LLM architectures.}
    \label{fig:averge}
\end{figure}

Fig.~\ref{fig:main_experiment} demonstrates that on the same dataset, the LLM fine-tuned by either AP-1k-\OURS~or EI-1k-\OURS~consistently outperforms the same LLM fine-tuned by baseline methods.
Notably, our method achieves an average win rate of 54.4\% across all the five testing sets compared to the current state-of-the-art method Longest~\cite{longest}, which only has a failure rate of 24\%~(see Fig.~\ref{fig:averge} with LLaMA2-7B). Moreover, EI-1k-\OURS~significantly outperforms LIMA-1k and the full Evol-Instruct-70k data, and shows consistent advantages over EI-1k-AG. Additionally, our score analysis indicates that the data from Evol-Instruct-70k generally receives higher scores, suggesting that Evol-Instruct-70k contains more high-quality data compared to Alpaca-52k.

\subsection{AlpacaEval 2.0 Evaluation}
We report the results on the AlpacaEval 2.0 benchmark of our models and baseline methods whose results are from the public leaderboard~\footnote{\url{https://tatsu-lab.github.io/alpaca_eval}}.
As shown in Table~\ref{tab:model_comparison}, \OURS~ consistently surpasses or stays on-par with SOTA methods. For instance, on Llama-2-7B, \textit{Refined-Alpaca-1k-\OURS} attains an 8.12\% win rate and surpasses several baselines that use far more data. When combined with NEFTune~\cite{neftune} that further augments IFT data with uniform noise, our win rate rises to 9.64\%, outpacing Tulu-2-DPO-7B (8.20\%)~\cite{tulu2} that utilizes 326k instruction-following samples and 64k preference pairs. Moreover, on the Mistral-7B-v0.1 architecture,~\OURS~reaches up to 13.77\% win rate with NEFTune, outperforming comparable methods that are trained on 10k instructions. A similar trend is also observed on Llama-2-13B, where our win rate increases to 11.35\% once NEFTune is enabled. Notably, expanding the generation limit from 2,048 to 4,096 tokens often leads to longer responses, yet does not necessarily produce higher win rates, hinting at a nuanced interplay between sequence length and model alignment. Overall, these findings underscore the effectiveness of our introspection-refinement approach, even at small data scales, and highlight the additional gains provided by NEFTune in boosting alignment performance.

\begin{table}[htbp]
\centering
\resizebox{0.5\textwidth}{!}{
\begin{tabular}{lcccc}
\toprule % 顶部加粗线
\textbf{Models} & \textbf{\# IFT Data} & \textbf{\# Pref. Data} & \textbf{Win Rate} & \textbf{Avg. Length} \\
\midrule % 第一条中间线加粗
\multicolumn{5}{l}{\textbf{Notable baselines}} \\
\toprule % 重新加粗\toprule[1.5pt] 
GPT-4-Turbo* & - & - & 50.0 & 2049 \\
Alpaca-7B* & 52k & 0 & 2.59 & 396 \\
Vicuna-7B* & 70k & 0 & 4.16 & 1044 \\
\hline
\multicolumn{5}{l}{Base model: \textbf{Llama-2-7B}} \\
\toprule % 加粗
Llama-2-Chat-7B* & 27k & 3M & 4.96 & 1479 \\
\hspace{1em}+ Evol70k-NEFTune* & 97k & 3M & 7.60 & 1612 \\
Tulu-2-DPO-7B* & 326k & 64k & 8.20 & 1663 \\
\hdashline % 第6行加虚线
AlpaGasus-1k & 1k & 0 & 2.69 & 745 \\
LIMA-1k & 1k & 0 & 2.74 & 1360 \\
Alpaca-52k & 52k & 0 & 2.74 & 586 \\
Alpaca-1k-longest & 1k & 0 & 3.16 & 1810 \\
\hspace{1em}+ max gen. 2048 $\rightarrow$ 4096 & 1k & 0 & 3.11 & 2290 \\
Alpaca-1k-\OURS$^{\square}$ & 1k & 0 & \textbf{6.21} & 1563 \\
\hspace{1em}+ max gen. 2048 $\rightarrow$ 4096 & 1k & 0 & 5.78 & 2045 \\
\hdashline
Evol-Instruct-70k & 70k & 0 & 3.44 & 850 \\
Evol-Instruct-1k-longest & 1k & 0 & 4.09 & 1866 \\
\hspace{1em}+ max gen. 2048 $\rightarrow$ 4096 & 1k & 0 & 4.16 & 2486 \\
Evol-Instruct-1k-\OURS$^{\square}$ & 1k & 0 & \textbf{7.11} & 1363 \\
\hspace{1em}+ max gen. 2048 $\rightarrow$ 4096 & 1k & 0 & 6.48 & 1759 \\
Evol-Instruct-AlpaGasus-1k & 1k & 0 & 4.32 & 1156 \\
\hdashline
Refined-Evol-Instruct-1k-longest & 1k & 0 & 5.12 & 1289 \\
Refined-Evol-Instruct-1k-\OURS$^{\square}$ & 1k & 0 & 7.96 & 1467 \\
Refined-Alpaca-1k-longest & 1k & 0 & 6.00 & 1732 \\
Refined-Alpaca-1k-\OURS$^{\square}$ & 1k & 0 & 8.12 & 1643 \\
\hspace{1em}+ max gen. 2048 $\rightarrow$ 4096 & 1k & 0 & 8.26 & 2147 \\
\hspace{1em}+ NEFTune & 1k & 0 & 9.64 & 1904 \\
\hspace{1em}+ NEFTune + 2048 $\rightarrow$ 4096 & 1k & 0 & \textbf{9.19} & 2072 \\
\hline
\multicolumn{5}{l}{Base model: \textbf{Mistral-7B-v0.1}} \\
\toprule % 加粗
Alpaca-52k & 52k & 0 & 3.42 & 450 \\
AlpaGasus-1k & 1k & 0 & 4.91 & 502 \\
LIMA-1k & 1k & 0 & 6.76 & 1197 \\
Alpaca-1k-longest & 1k & 0 & 7.13 & 937 \\
Alpaca-1k-\OURS$^{\square}$ & 1k & 0 & 10.55 & 1130 \\
Refined-Alpaca-1k-longest & 1k & 0 & 11.74 & 1170 \\
Refined-Alpaca-1k-\OURS$^{\square}$ & 1k & 0 & 13.29 & 1348 \\
\hspace{1em}+ max gen. 2048 $\rightarrow$ 4096 & 1k & 0 & 13.45 & 1593 \\
\hspace{1em}+ NEFTune & 1k & 0 & \textbf{13.77} & 1370 \\
\hline
\multicolumn{5}{l}{Base model: \textbf{Llama-2-13B}} \\
\toprule % 加粗
Alpaca-52k & 52k & 0 & 3.90 & 556 \\
Alpaca-1k-longest & 1k & 0 & 4.80 & 1104 \\
Alpaca-1k-\OURS$^{\square}$ & 1k & 0 & 7.28 & 864 \\
AlpaGasus-1k & 1k & 0 & 4.87 & 540 \\
LIMA-1k & 1k & 0 & 5.64 & 1097 \\
Refined-Alpaca-1k-longest & 1k & 0 & 8.44 & 1646 \\
Refined-Alpaca-1k-\OURS$^{\square}$ & 1k & 0 & 10.72 & 1298 \\
\hspace{1em}+ max gen. 2048 $\rightarrow$ 4096 & 1k & 0 & 10.65 & 1539 \\
\hspace{1em}+ NEFTune & 1k & 0 & \textbf{11.35} & 1194 \\
\bottomrule % 加粗
\end{tabular}
}
\caption{\textbf{Preference Evaluation Results on AlpacaEval 2.0.} The evaluation considers factors such as training data, win rates, and average output lengths. For our models, unless otherwise specified, we use a generation limit of 2048 tokens. * denotes results from the leaderboard. $^{\square}$ indicates models developed by us, and bold numbers highlight the best performance within the same dataset.}
\label{tab:model_comparison}
\end{table}

\subsection{Evaluation on Instruction-Following}
To better demonstrate our \OURS's superiority over others, we conduct experiments on MT-Bench~\cite{mt-bench} to evaluate its instruction-following capability. As shown in Tab.~\ref{tab:example}, using LLaMA2-7B as the base model, unrefined AP-1k-\OURS~significantly outperforms all the other unrefined baselines. Notably, unrefined AP-1k-\OURS~achieves on-par performance with Refined-AP-1k-Longest. Moreover, when incorporating the refinement method from~\cite{longest}, which leverages GPT-4~\cite{achiam2023gpt} for instruction refinement, our method attains a score of \textbf{5.77}, surpassing others by a considerable margin.
When the base LLM is changed to LLaMA2-13B or Mistral-7B, our approach still shows consistent improvements over baseline methods, which further confirms our generalizability and effectiveness across different LLM architectures.

\begin{table}[htbp]
    \centering
    \caption{\textbf{Single-Score Evaluation on MT-Bench~\cite{mt-bench} across different base LLMs and IFT datasets.} Scores are generated by GPT-4 with a range of [1, 10]. The best performing methods are highlighted as bold, while the second best ones are marked with underlines.}
    \begin{minipage}{0.49\textwidth} % Ensure the table stays within half a column
        \begin{tabular}{l@{\hskip 0.05in}c@{\hskip 0.05in}c@{\hskip 0.05in}c} % Adjust the column spacing
            \toprule
            \multirow{2}{*}[-2.2ex]{Datasets} & \multicolumn{3}{c}{Models} \\
            \cmidrule(l{0.5pt}r{0.5pt}){2-4}
            & LLaMA2-7B & LLaMA2-13B & Mistral-7B \\
            \midrule
            Alpaca-52k~\cite{StanfordAlpaca} & 3.74 & 5.40 & 5.35 \\
            LIMA-1k~\cite{Lima} & 3.95 & 5.18 & \textbf{6.18} \\
            AP-1k-AG~\cite{chen2024alpagasus} & 3.63 & 4.70 & 6.06 \\
            AP-1k-Longest~\cite{longest} & 3.96 & 5.32 & 5.80 \\
            \midrule
            AP-1k-\OURS~(\textit{Ours}) & 4.17 & 5.47 & 6.00 \\
            \midrule
            Refined-AP-1k-Longest~\cite{longest} & \underline{4.18} & \textbf{6.09} & 6.00 \\
            Refined-AP-1k-\OURS~(\textit{Ours}) & \textbf{5.77} & \underline{6.02} & \underline{6.08} \\
            \bottomrule
        \end{tabular}
    \end{minipage}
    \label{tab:example}
\end{table}

\subsection{Evaluation with Automatic Metrics}
To eliminate the potential biases introduced by the judge LLMs, we conduct additional experiments using automatic evaluation metrics including ROUGE~\cite{lin2004rouge} and BLEU~\cite{papineni2002bleu} scores. As shown in Tab.~\ref{tab:metrics_comparison},~\OURS~consistently outperforms baseline models on both metrics, confirming our significance for IFT data selection.  

\begin{table}[htbp]
    \centering
    \large
    \vspace{0.2cm}
    \caption{\textbf{Performance Comparison on MT-Bench~\cite{mt-bench} with GPT-4 as Reference Standard.} The table presents ROUGE and BLEU scores calculated by comparing generated responses with GPT-4's answers. The best-performing results are highlighted in bold.}
    \label{tab:metrics_comparison}
    \resizebox{0.998\linewidth}{!}{
    \renewcommand{\arraystretch}{0.6}
    \begin{tabular}{l | c c c c }
    \toprule
    \textbf{Datasets} & \textbf{ROUGE-1} & \textbf{ROUGE-2} & \textbf{ROUGE-L} & \textbf{BLEU} \\ 
    \midrule
    AP-1k-\OURS~(Ours)    & \textbf{0.284} & \textbf{0.104} & \textbf{0.243} & \textbf{0.088} \\ 
    Alpaca-52k~\cite{StanfordAlpaca}    & 0.229          & 0.086          & 0.187          & 0.026          \\ 
    AP-1k-AG~\cite{chen2024alpagasus}      & 0.186          & 0.076          & 0.166          & 0.050          \\ 
    AP-1k-Longest~\cite{longest} & 0.261          & 0.102          & 0.199          & 0.026          \\ 
    \bottomrule
    \end{tabular}
    }
\end{table}

\section{ABLATION STUDIES}
To provide a better understanding of \OURS's design choices, we conduct ablation experiments on the LLaMA2-7B model with the Alpaca-52k dataset. As shown in Tab.~\ref{tab:ablation_results}, four settings are designed to reveal the influence of Open Tagging (OT), Comparative Scoring (CS), instruction-based embedding, and different score ranges, respectively. Results and analysis are detailed below.

\noindent \textbf{(1) Open Tagging (OT).}
We first assess the role of OT by constructing a variant without OT (\OURS{} w/o OT in Tab.~\ref{tab:ablation_results}). When compared with this variant, the full~\OURS~model achieves 1,310 wins, 170 ties, and 580 losses, resulting in a win rate of 1.354. This ablation study indicates that OT plays a pivotal role in IFT data selection, ultimately leading to better model performance.

\noindent \textbf{(2) Comparative Scoring (CS).} We then validate the importance of CS by replacing the pairwise comparative mechanism with an individual scoring strategy (\OURS{} w/o CS), which results in 1,345 wins and a win rate of 1.388. This further confirms that comparing samples through pairwise scoring helps the judge model distinguish subtle quality differences and select high-quality data for IFT.

\noindent \textbf{(3) Instruction-based Embedding (IE).} We further explore the scenario in which data samples are embedded using only their original instructions (i.e., \OURS{} w/o Tagging [1, 100]). Although directly using raw labels yields some performance, it is noticeably lower than the fully equipped~\OURS, which achieves a 1.375 win rate. These results suggest that decomposing instructions into tag features offers richer semantic granularity, boosting the selection effectiveness.

\noindent \textbf{(4) Original [1, 10] Score Range.} 
Finally, we revert the scoring range to [1, 10] while keeping all other modules intact. Under this configuration,~\OURS~achieves 1,586 wins, 275 ties, and 199 losses, yielding a win rate of 1.673. However, this setting underperforms compared with the broader range, indicating that a more fine-grained scoring scheme yields greater discrimination of sample quality and ultimately better guides the selection of fine-tuning data.

\begin{table}[htbp]
    \centering
    \caption{\textbf{Ablation Studies with LLaMA2-7B and Alpaca-50k.} Results are averaged on five test sets, comparing~\OURS~with variant models.
    Win rate = (\#Win $-$ \#Lose) / (Total) + 1.}
    \label{tab:ablation_results}
    \begin{tabular}{l@{\hskip 8pt}c@{\hskip 5pt}c@{\hskip 5pt}c@{\hskip 5pt}c@{\hskip 5pt}c}
        \toprule
        Original~\OURS~\\vs. Variant Models & Score Range & \#Win & \#Tie & \#Lose & Win Rate \\
        \midrule
        \OURS{} w/o OT & [1, 100] & 1,310 & 170 & 580 & \textbf{1.354} \\
        \OURS{} w/o CS & [1, 100] & 1,345 & 169 & 546 & \textbf{1.388} \\
        \OURS{} w/ IE & [1, 100]  & 1,175 & 482 & 403 & \textbf{1.375} \\
        \OURS{} & [1, 10]  & 1586 & 275 & 199 & \textbf{1.673} \\
        \bottomrule
    \end{tabular}
\end{table}

% \subsection{Ablation Study on Tag Features and Score Range Expansion}
% we evaluate the impact of three critical design choices in our framework: (1) the use of tag features, and (2) the expansion of score ranges.

% \noindent\textbf{Tag Features.} Our tagging mechanism decouples the complex semantic meaning of each IFT sample into multiple open-domain tags. This fine-grained tagging enables more accurate clustering and rating of IFT samples, allowing multiple clusters to collaboratively evaluate each sample, thus improving data diversity and selection quality.

% \noindent\textbf{Score Range Expansion.} Initial experiments with the original score range showed an overabundance of tie cases, making it difficult to distinguish salient IFT samples. To address this, we expanded the score range and set a refined criterion, resulting in a more differentiated scoring mechanism that better captures the differences between high-quality and low-quality samples.

% The ablation studies on (1) and (2) are shown in Tab.~\ref{tab:ablation_2}, and the impact of paired scoring can be found in the last row of Tab.~\ref{tab:ablation_results}. Collectively, these design choices enhance the selection of salient IFT samples, ultimately improving the overall model performance.

\section{CONCLUSION}
In this work, we presented~\OURS, a novel approach that integrates Open Tagging and Comparative Scoring to improve data diversity and scoring reliability in IFT data selection, respectively. More specifically, the Open Tagging mechanism maintains data diversity through autonomous open-domain tags assigned by LLMs and reduces tag redundancy via normalization that leads to more efficient clustering. The Comparative Scoring paradigm first adopts fine-grained prompts for better alignment with human evaluation standards and then evaluate data samples in a pairwise comparative fashion, thereby keeping evaluation criteria consistent across data and improving score reliability. 
Extensive experiments on diverse datasets and LLM architectures have been conducted to validate our proposed method. Results demonstrate that~\OURS~significantly enhances the effectiveness of IFT data selection, offering a promising solution for optimizing LLM performance with IFT.

% \newpage
% \cleardoublepage  % 如果你希望它总是在奇数页开始（对于双面打印很有用）
\bibliographystyle{IEEEbib}
\bibliography{icme2025references}
\end{document}